\documentclass[conference]{IEEEtran}
\IEEEoverridecommandlockouts
\usepackage{cite}
\usepackage{amsmath,amssymb,amsfonts}
\usepackage{algorithmic}
\usepackage{graphicx}
\usepackage{textcomp}
\usepackage{xcolor}
\usepackage{dblfloatfix}
\usepackage{float}
\usepackage{balance}
\def\BibTeX{{\rm B\kern-.05em{\sc i\kern-.025em b}\kern-.08em
    T\kern-.1667em\lower.7ex\hbox{E}\kern-.125emX}}
\begin{document}

\title{Octopus-Swimming-Like Robot with Soft Asymmetric Arms\\
}

\author{Bobing Zhang$^{\dagger}$, Yiyuan Zhang$^{\dagger}$$^{*}$,~\IEEEmembership{Graduate Student Member,~IEEE}, Yiming Li, Sicheng Xuan, Hong Wei Ng, \\Yuliang Liufu, Zhiqiang Tang,~\IEEEmembership{Member,~IEEE}, Cecilia Laschi,~\IEEEmembership{Fellow,~IEEE}

\thanks{$^{\dagger}$These authors contributed equally to this work.

$^{*}$Corresponding author: Yiyuan Zhang (yiyuan.zhang@u.nus.edu)

Department of Mechanical Engineering, National University of Singapore, Singapore 117575, Singapore.

This work was supported by following projects: the DESTRO project (``Dextrous, strong yet soft robots,'' R22I0IR124), funded by MAE (Italy) and A*STAR (Singapore); the REBOT project (``Rethinking underwater robot manipulation,'' Moe-t2eP50221-0010), funded by the Singapore Ministry of Education; and the RoboLife project (``Soft robots with morphological adaptation and life-like abilities''), funded by the NUS Funding Agency (Singapore).
}
}

\maketitle

\begin{abstract}
 
Underwater vehicles have seen significant development over the past seventy years. However, bio-inspired propulsion robots are still in their early stages and require greater interdisciplinary collaboration between biologists and roboticists. The octopus, one of the most intelligent marine animals, exhibits remarkable abilities such as camouflaging, exploring, and hunting while swimming with its arms. Although bio-inspired robotics researchers have aimed to replicate these abilities, the complexity of designing an eight-arm bionic swimming platform has posed challenges from the beginning. In this work, we propose a novel bionic robot swimming platform that combines asymmetric passive morphing arms with an umbrella-like quick-return mechanism. Using only two simple constant-speed motors, this design achieves efficient swimming by replicating octopus-like arm movements and stroke time ratios. The robot reached a peak speed of 314 mm/s during its second power stroke. This design reduces the complexity of traditional octopus-like swimming robot actuation systems while maintaining good swimming performance. It offers a more achievable and efficient platform for biologists and roboticists conducting more profound octopus-inspired robotic and biological studies.

\end{abstract}


\section{Introduction}
The Earth is predominantly covered by water, sparking significant human interest in exploring and understanding underwater environments. To reduce risks and save costs, underwater robots are increasingly replacing humans in various underwater tasks \cite{b1}. A significant category of these robots includes underwater vehicles. In addition to traditional remotely operated vehicles (ROVs), bio-inspired robots have gained considerable attention in recent research. Marine animals have evolved to navigate aquatic environments efficiently, utilizing propulsion mechanisms that are both interactive and adaptive \cite{b2}. Compared to conventional fuel-powered underwater vehicles, bio-inspired designs not only maintain propulsion efficiency but also reduce noise pollution, minimizing the environmental impact on marine ecosystems \cite{b3}.

The octopus, an underwater mollusk, is a classic example of biomimetic robotics (Fig. 1) \cite{b4}. The octopus's body has no bones or rigid components, meaning it has unlimited freedom of movement. In addition, the good movement ability and intelligent behavior of octopuses have also attracted significant attention from many roboticists and biologists \cite{b5}. M Calisti et al. proposed that the counterforce generated by the contact between octopus arms and the ground drives the octopus to move forward \cite{b6}. M Sfakiotakis et al. developed an octopus swimming robot consisting of eight octopus arms made of polyurethane to create instant thrust by controlling the time ratio of the opening and closing of the eight legs \cite{b7}. However, each octopus arm of these robots is driven by a single motor. A large number of motors will lead to higher complexity and more fault points. The probability of failure will significantly increase under conditions requiring high dexterity \cite{b8}. According to biological octopus arm swimming observation, the recovery stroke takes longer than the power stroke in one stroke period (see Fig. 2) \cite{b7}, which means that the octopus spends more time opening its arms than closing it to propulsion fast.

\begin{figure}
\centerline{\includegraphics[width=\columnwidth]{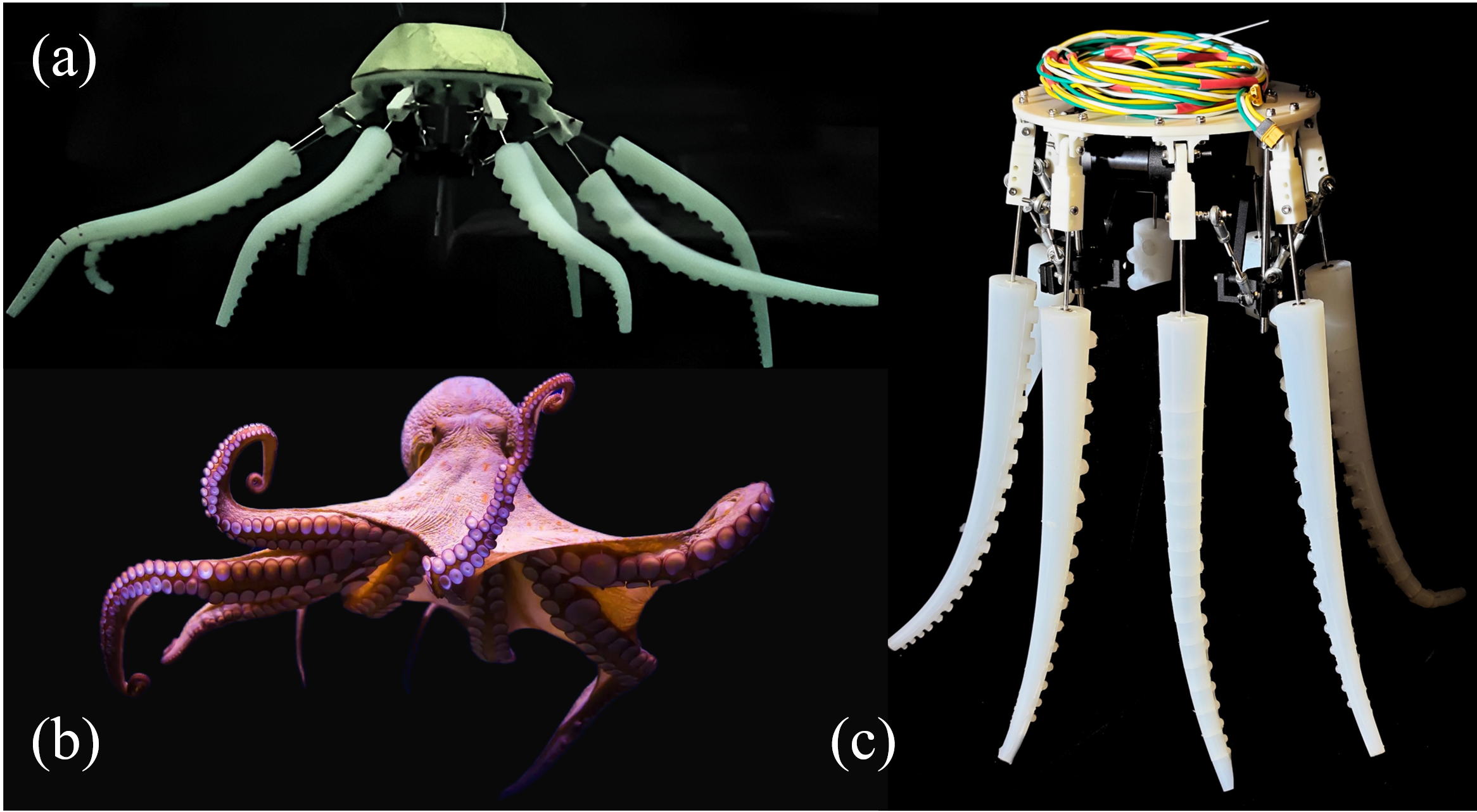}}
\vspace{-2mm}
\caption{Overview of our octopus swimming robot. a) Our octopus robot swimming in water. b) Biological octopus swimming in water. Image credit: TheSP4N1SH/iStock. c) Octopus swimming robot in this project.}
\label{Fig. 1}
\vspace{-6mm}
\end{figure}

Jan et al. designed a fully soft octopus robot based on the principle of fluidic actuation \cite{b9}. The eight arms, each containing fluid-filled drive chambers, were actuated by a continuous fluid system that deformed them in various directions to generate forward thrust. Faheem et al. proposed an octopus robot that contains a pair of shape memory alloy (SMA) muscle wires in each arm. The octopus arm was bent by controlling the SMA on the arm to obtain forward thrust \cite{b10}. This type of robot reduces the number of actuators. However, the practical SMA actuator control is still challenging for the field \cite{b11}. In the biological swimming stroke period, the octopus arm shape morphs. In a recovery stroke, the arms slowly and softly open to be ready, as shown in Fig. 2 (a). Then come the power strokes, during which the octopus arms deform from curved to straight states\cite{b12}, shown in Fig. 2 (b). Clemente et al. proposed that the muscle in the octopus arm is mobilized to participate in stiffening the arms to resist the deformation caused by water \cite{b13}. In the current soft robot research, similar stiffness adjustability needs a complex actuation and control system. Most swimming octopus robots ignore the stiffening feature in the swimming process. They have the same stiffness when doing recovery and power strokes. 

\begin{figure}
\centerline{\includegraphics[width=\columnwidth]{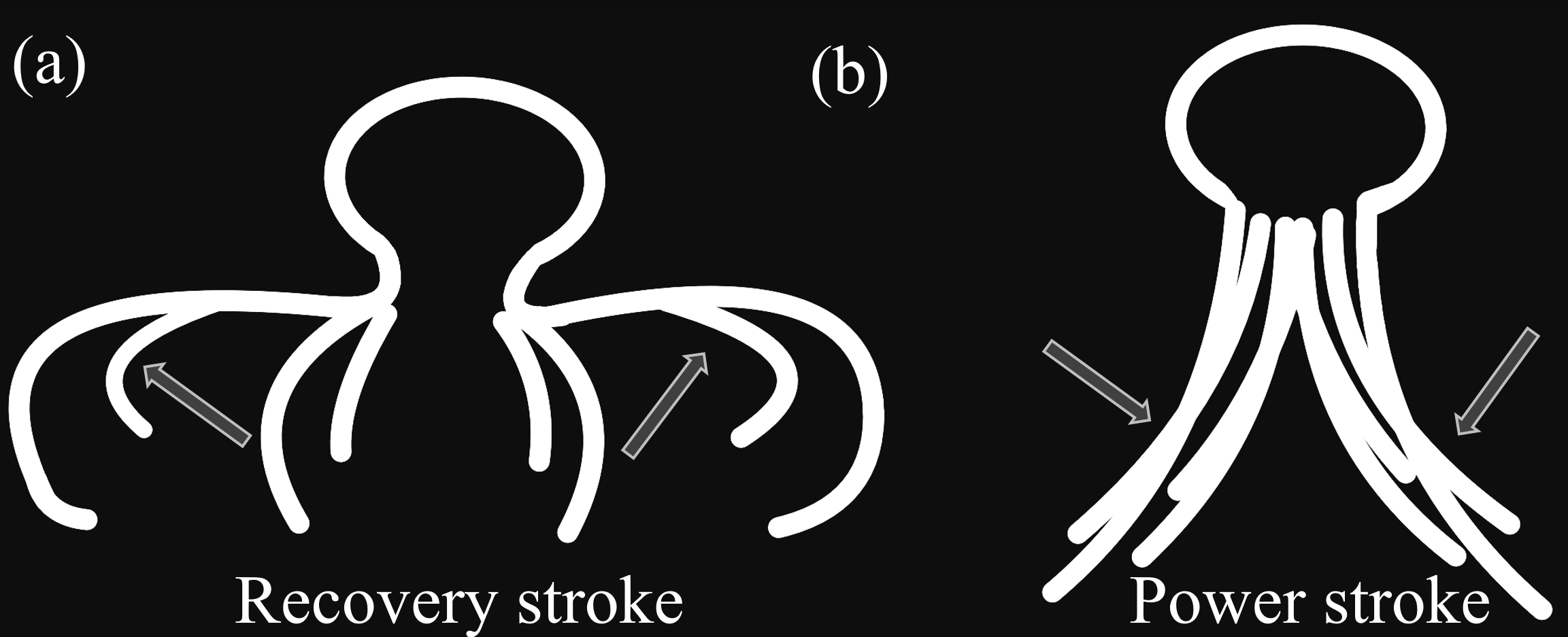}}
\vspace{-2mm}
\caption{Recovery stroke and power stroke of octopus. a) The recovery stroke describes the process of the octopus arms moving from a closed to an open position. b) The power stroke describes the process of the octopus arms moving from an open to a closed position.}
\label{Fig. 2}
\vspace{-5mm}
\end{figure}

This project designed an umbrella-like mechanism that enabled the repeatable full-stroke process of swimming with only two motors. This design dramatically reduces the actuation complexity of the octopus swimming robot. Additionally, the mechanism can adjust the time ratio between the recovery stroke and power stroke to 2.0:1, closely resembling the movement of a biological octopus \cite{b14}. Besides, the newly proposed asymmetrical octopus arm passively behaves with different stiffness in the recovery and power stroke process without any additional control. This design allows the soft artificial arms to closely mimic the swimming trajectory of a biological octopus arm.

\section{Method}
The octopus robot integrates an umbrella-like stroke mechanism with asymmetrically stiff octopus arms. It consists of eight legs, each made of Dragon Skin-30 silicone by injection moulding. After moulding, the arm has been partially cut to a certain depth to achieve different asymmetric bend stiffness. The umbrella-like opening and closing mechanism is assembled with 3D-printed parts and two motors. Fig. 1 (b) to Fig. 1 (d) shows the 3D model of the umbrella-like mechanism.

\subsection{Design of Mechanism}

When an octopus swims, the opening and closing movements of its arms are similar to the motion of an umbrella opening and closing. Capturing this inspiration, an umbrella-like opening and closing mechanism is proposed. As Fig.\nobreakspace 1 (b) shows, the connecting rod is designed to install the artificial octopus arm. One end is connected to the chassis of the octopus robot head, and the other is connected to the slider through the support rod. As we mentioned, the time ratio between the recovery stroke process and the power stroke process in octopus swimming is about 2.0:1. To simulate the time ratio of a biological octopus, the design incorporates an offset crank slider mechanism common in engineering. The crank slider mechanism converts the reciprocating motion of the slider into rotation motion of the crank. For a specific stroke time ratio, only a relatively constant speed motor rotation is needed. Four sets of connecting rods and support rods are assembled on the sliding block like half an umbrella to replicate the four swimming arms of an octopus. The entire umbrella-like mechanism consists of two sets of four arms, each controlled by a motor. This configuration enables the octopus robot to execute a series of turns by sending different control commands to the two motors.

\begin{figure}
\centerline{\includegraphics[width=0.9\columnwidth]{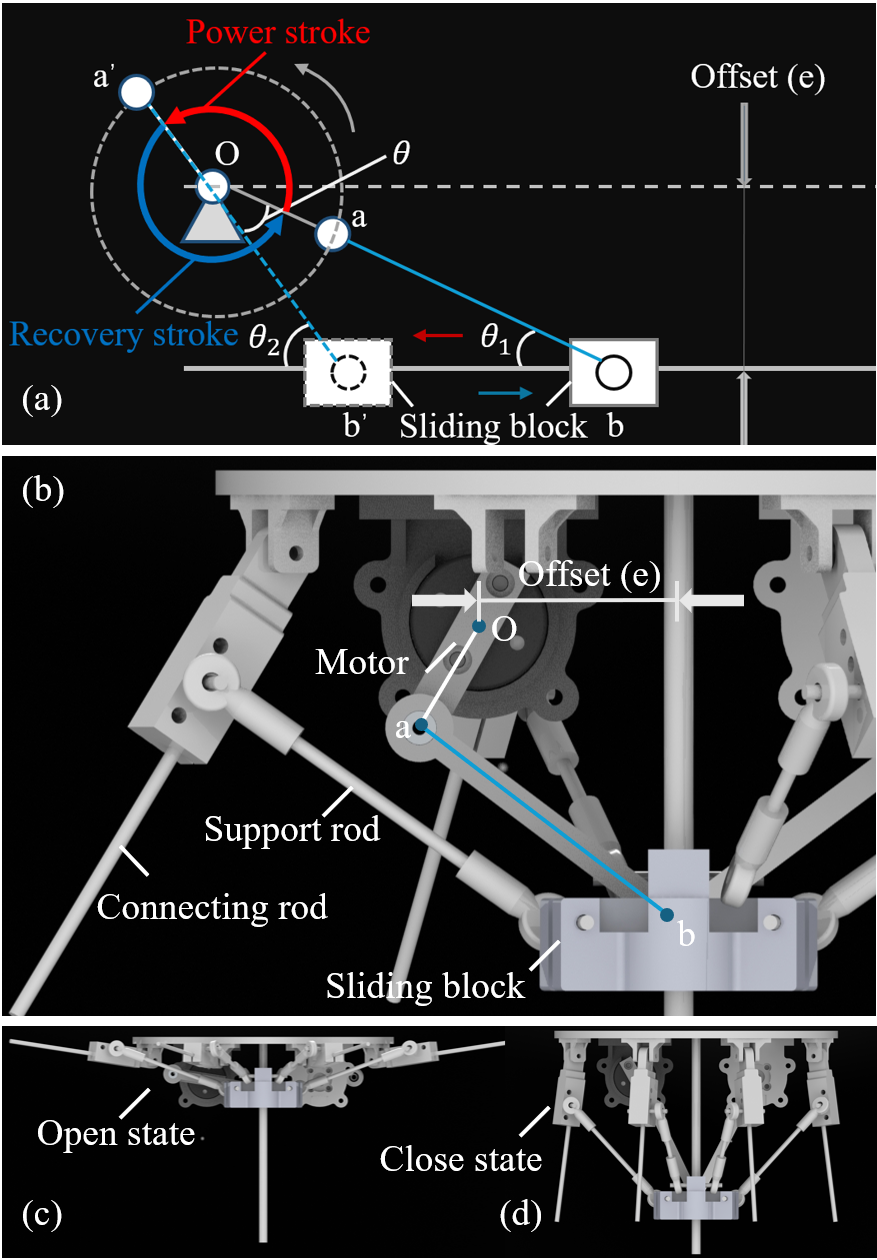}}
\vspace{-2mm}
\caption{The application of the offset crank slider mechanism in this design. a) Schematic diagram of offset crank slider. b) The offset crank slider mechanism can be fused into the octopus mechanism. c) The open state of the mechanism corresponds to the final position of the recovery stroke process. d) The closed state of the mechanism corresponds to the final position of the power stroke process.}
\label{Fig. 3}
\vspace{-6mm}
\end{figure}

\begin{figure*}[!b]
\centerline{\includegraphics[width=\textwidth]{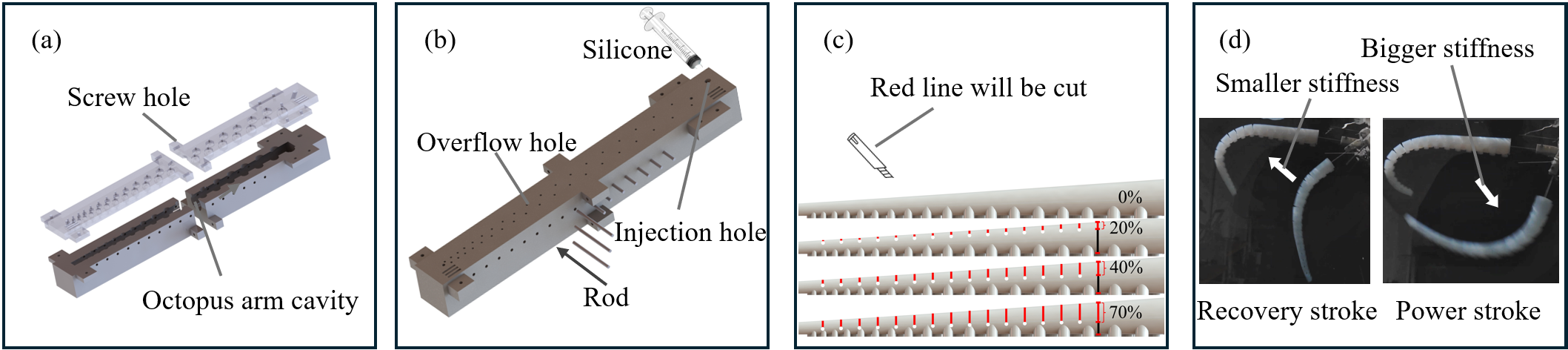}}
\vspace{-2mm}
\caption{A flowchart for making an octopus arm. a) The mould is printed from polylactide (PLA) material. b) Injection moulding is used to make silicone octopus arms. c) Different depths of incisions are cut along the back of the octopus arms using a knife. d) Octopus arms have less stiffness in recovery stroke process and more stiffness in power stroke process.}
\label{Fig. 4}
\vspace{-6mm}
\end{figure*}

As shown in Fig. 3 (a), the offset crank slider mainly comprises three parts: crank Oa, connecting rod ab, and slider b. Offset (e) indicates the distance between the track where the slider is located and the rotation center O. The center line of the guide rail of the slider does not pass through the rotation center O of the crank Oa. When the mechanism runs, the crank Oa is used as a rotating rod, carrying out uniform circular motion around the O point. The connecting rod ab acts as a slave rod, and the other end is connected to the slider. When the two rods overlap on the same line (at Oa’ and a’b), the slider’s position is at the leftmost end of the guide rail. Then, the Oa rod starts to rotate counterclockwise as the recovery stroke process as blue circular arrow shows, and the slider starts to move to the right end of the guide rail. When the two rods move to the same line again on the other side of O (at Oa and ab), the slider reaches the rightmost end of the guide rail. When the rod is turned counterclockwise again as the power stroke process as red circular arrow shows, the slider begins to move to the left end of the guide rail. When the two rods coincide again, the slider returns to the leftmost end, completing a cycle of operation. As the crank Oa rotates at a constant speed, the rotating angle ratio of the power stroke and recovery stroke process determines the time ratio of the reciprocating motion of the slider. The angle $\theta$ between these two limit positions (Ob’ and Ob) is called the polar angle. When the slider is used as the output part of the mechanism, the mechanism has a quick return characteristic. 

Fig. 3 (b) shows the mechanism gives our octopus robot a quick-return feature. The motor center is expressed as the O point, and the horizontal distance between the O point and the slider is expressed as Offset distance e in the offset crank slider. Bearings are used between the rods to minimize friction. The motor, crank Oa, and connecting rod ab work together to control the up and down movement of the slider. As shown in Fig. 3 (c) and (d), the slider rises to lift the connecting rods (which attach to the artificial arms) up and falls to lower them down. The time ratio between the power stroke and recovery stroke processes can be controlled by adjusting the lengths of the Oa rod and ab rod. The ratio of the average angular velocity between the rising and falling of the slider can be represented by the travel ratio coefficient $K$, which indicates the degree of the quick-return characteristic of the biased crank-slider mechanism. 

Assuming that the angles of a general biased crank-slider mechanism in the push and return are $\phi_1$ and $\phi_2$, since the mechanism rotates at a uniform speed, the coefficient of travel velocity ratio can be expressed as:

\begin{equation}
    K=\frac{w_1}{w_2}=\frac{\phi_1/t_1}{\phi_2/t_2}=\frac{\phi_1}{\phi_2}=\frac{180^\circ+\theta}{180^\circ-\theta}
\end{equation}

\begin{equation}
    \theta=180^\circ(K-1)/(K+1)
\end{equation}

The angle between the critical points can be calculated after determining the $K$ value.The angle between the critical points can also be calculated by the Pythagorean theorem.

\begin{equation}
    \theta_1=arcsin(\frac{e}{a+b})
\end{equation}

\begin{equation}
    \theta_2=arcsin(\frac{e}{b-a})
\end{equation}

\begin{equation}
    \theta=\theta_1-\theta_2
\end{equation}

The lengths of the Oa and ab rods can be designed according to the desired travel ratio coefficient. To study the effects of different recovery and power stroke time ratios on the swimming speed of the octopus robot, three time-ratio mechanisms were developed: 2.0:1, 1.6:1, and 1.2:1. The octopus swimming robot features a head chassis with a diameter of 190 mm and an offset of 40 mm. For the 2.0:1 stroke time ratio, the crank and slider lengths are 25 mm and 66 mm, respectively. For the 1.6:1 ratio, the lengths are 25 mm and 69.4 mm, while for the 1.2:1 ratio, the lengths are 19.5 mm and 83 mm.

\subsection{Soft Asymmetric Arms}

When an octopus swims upward in the water, it first moves all its arms up to the height of its head and then moves its arms down to make a power stroke. Kawk mentioned that when the swimming paddle of the swimming robot has greater rigidity, it can generate more thrust, especially at higher swimming frequencies \cite{b15}. Since the primary objective of the octopus robot is to achieve upward movement, it is essential to minimize the unwanted thrust generated during the recovery stroke of the octopus arm while maximizing the thrust produced during the power stroke. To achieve this, octopus arms with asymmetrical stiffness are created. Fig. 4 shows the manufacturing process of the asymmetric stiffness octopus arms. Fig. 4 (a) illustrates the mould used to create the octopus arm, featuring screw holes that allow the four mould parts to be tightly fastened together. Once assembled, the enclosed area of the mould reveals the overall shape of the octopus arm. Fig. 4 (b) depicts the injection moulding process. To prevent incisions in the octopus arm from causing cracks or stress concentrations that could damage the arm, aluminium rods are inserted into the mould to create small, evenly distributed holes in the arm. After the mould is fully assembled, silicone is injected until it flows out of the overflow holes. Subsequently, the entire mould is placed in an oven at 40 ºC for 40 minutes. Fig. 4 (c) shows how the moulded arm will be cut to varying depths to achieve different asymmetric mechanical properties.

As shown in Fig. 4 (d), during the recovery stroke of the artificial octopus arm, it moves upward and bends inward passively due to the effect of water drag, facilitated by its relatively lower directional bending stiffness. As a result, the gaps between discrete silicone protrusions can be observed. This passive deformation may help further reduce water drag, minimizing thrust in undesired directions. When the artificial octopus arm is in the power stroke process, the octopus arm moves down with slight passive bending. At this point, the discrete silicone protrusions squeeze into each other tightly as an arm without gaps. The octopus arm faces water drag with the complete octopus arm bend stiffness. To study how incision depth on the back of the octopus arms makes the artificial arm trajectory most similar to the biological octopus arms, we set up four sets of artificial octopus arms with different depths. These octopus arms had 0\%, 20\%, 40\%, and 70\% depth incisions. As shown in Fig. 4 (c), there is no other difference between the four arms groups except the incision depth. All arms are 300 mm in length, 30 mm in diameter at the base, and 10 mm in diameter at the tip.

Clements mentioned that octopuses have the characteristic of neutral buoyancy in water \cite{b16}. To simulate the octopus swimming in the water, we added floating blocks to the whole mechanism so that the entire mechanism could be neutral buoyancy in the water.

\section{Results}

\subsection{Single Arm Swimming Behavior With Different Asymmetry}

\begin{figure}
\centerline{\includegraphics[width=\columnwidth]{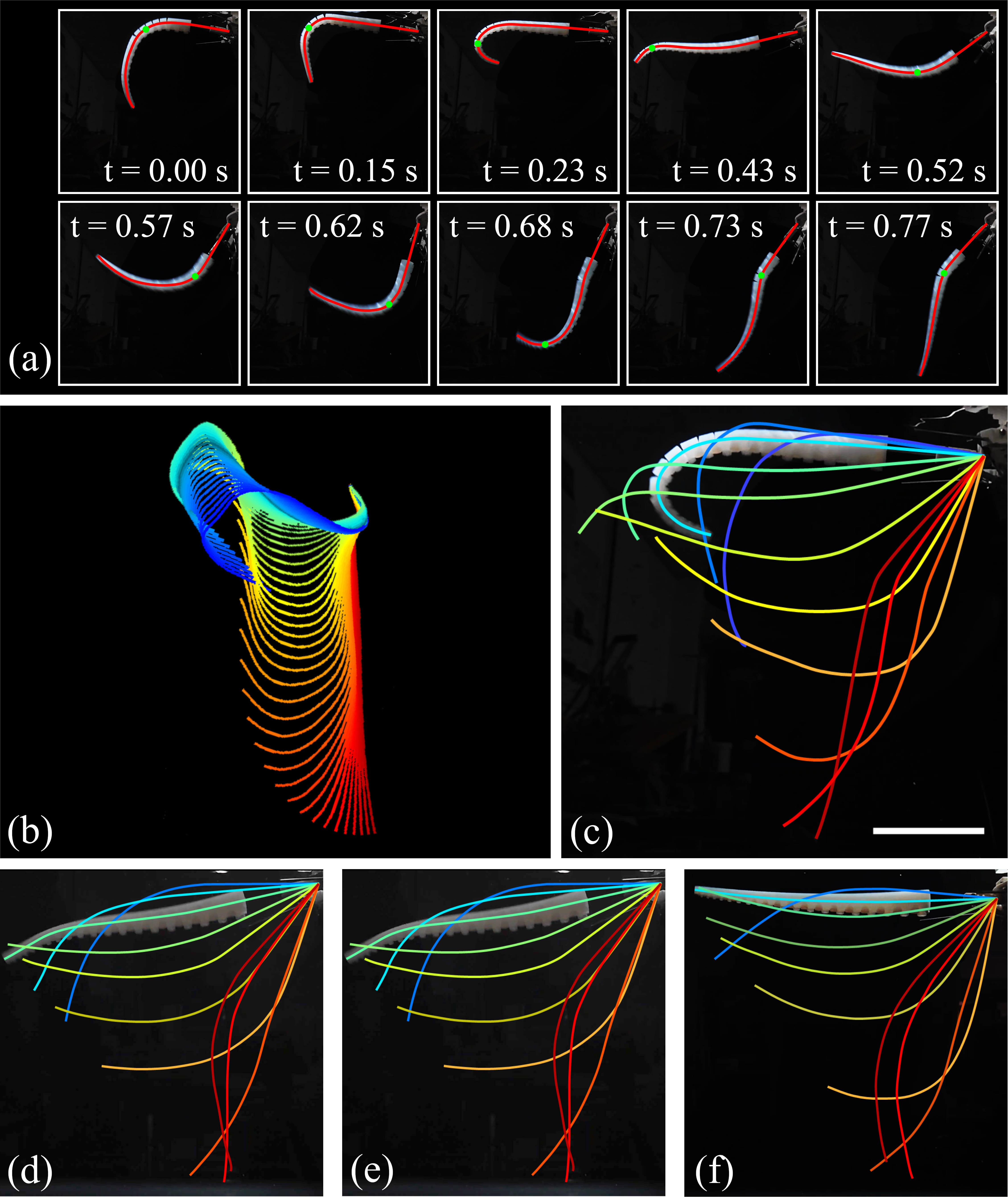}}
\vspace{-2mm}
\caption{Comparison between the stroke trajectory diagram of an asymmetric stiffness artificial arm and a biological octopus arm. a) The motion trajectory of the artificial arm over time with a cutting depth of 70\%. The green dot is where the arm has the largest curvature. b) The motion trajectory of biological octopus arm \cite{b12}. c) The motion trajectory of artificial arm with a cutting depth of 70\%. The white scale represents a length of 10 cm. d) The motion trajectory of artificial arm with a cutting depth of 40\%. e) The motion trajectory of artificial arm with a cutting depth of 20\%. f) The motion trajectory of the artificial arm with a cutting depth of 0\%.}
\label{Fig. 5}
\vspace{-6mm}
\end{figure}

\begin{figure*}
\centerline{\includegraphics[width=0.8\textwidth]{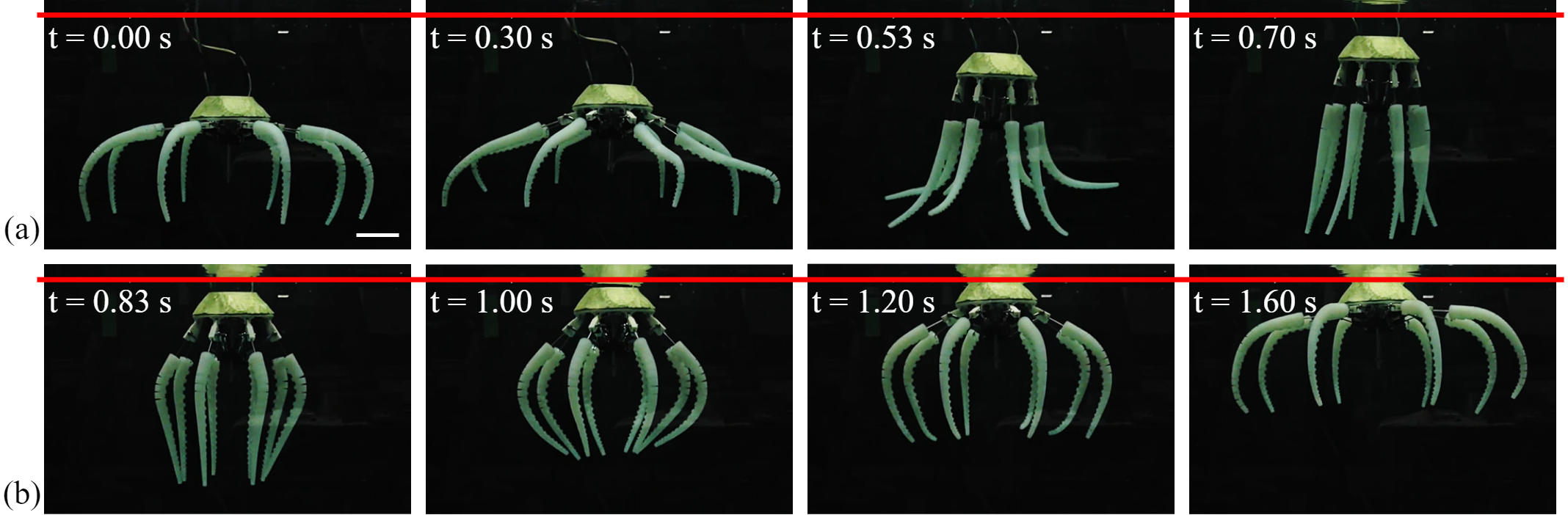}}
\vspace{-2mm}
\caption{Swimming schematic of octopus robot with a stroke time ratio of 2.0:1. a) The first row shows the power stroke process. b) The second row shows the recovery stroke process.}
\label{Fig. 6}
\vspace{-5mm}
\end{figure*}

During our research, we found that adding asymmetric structure to the robotic octopus arm helped to improve its motion trajectory. To test the effect of the octopus arm asymmetry on its swimming trajectory, we made octopus arms with different cutting depths of 0\%, 20\%, 40\%, and 70\%. All experiments were conducted at a motor speed of 48 revolutions per minute. The result of capturing the motion trajectory of the single octopus arm in water is shown in Fig. 5. In the paper by Asimina Kazakidi et al. \cite{b12}, the posture of the single-arm swimming motion of biological octopuses is discussed. They plotted the positions generated by the arm movement in chronological order (Fig. 5 (b)). The color gradient represents the progression of time. The more the color shifts towards blue, the earlier it occurs in the entire movement process. Throughout the whole movement period, the maximum curvature of the midline first increases and then decreases with orientation change. Notably, in the early phase of the movement (dark blue curve), the curve points back to the base of the arm, which we call a ``recurve'' at the end of the arm.

In the whole operation cycle of each asymmetric artificial arm, ten tracks corresponding to the trajectory data of the biological octopus arm were extracted. By capturing frames at different times, we analyzed the overall motion characteristics of the arm with a cutting depth of 70\% (see Fig. 5 (a)). The maximum curvature point first moved upwards and away from the arm base (near the drive end, on the right side of the picture), and the curvature gradually increased during the initial period before t = 0.23 s. The bending gradually decreased when the arm reached its highest point and became horizontal. Later, the curvature of the maximum curvature point was relatively minor when pushing water downwards. The maximum curvature point was also transferred from the base of the arm to the tip of the arm (away from the drive end, on the left side of the image). However, at this time, the bend was pointing downward. At t = 0.73 s, the arm reached the lowest point of the mechanism and started to move upwards. Between t = 0.73 s and t = 0.77 s, the maximum curvature orientation reverses. Also, as it approached the end of the movement period, the maximum curvature value became smaller. We also assigned the colours from blue to red in a gradient according to the sequence and plotted them on the same image (Fig. 5 (c)-(f)). 

The experiment demonstrated that only the artificial arm with a 70\% cutting depth showed a biological-like curvature at the distal end of the arm at the t = 0.23 s. This curvature was more evident than the curvature at t = 0.68 s, as shown in Fig. 5 (a). We can infer that asymmetric design has a significant role in passive interaction with water, which will have a positive effect on underwater swimming propulsion. As shown in Fig. 5 (b), Fig. 5 (d), Fig. 5 (e), and Fig. 5 (f), the motion trajectory of artificial arms with 40\%, 20\%, and 0\% cutting depth was substantially different from the motion trajectory of the biological octopus arm. We applied the 70\% cutting length arms for the following swimming experiment due to its good biological similarity.

\subsection{Robot Swimming Performance of Different Stroke Ratio}

To investigate how the stroke time ratio influences octopus swimming, we manufactured rapid return mechanisms with stroke time ratios of 2.0:1 (matching that of a biological octopus), 1.6:1, and 1.2:1 to evaluate their impact. We used a 1.5 m cubic tank with a 1.3 m water depth for the experiment. The experiments were conducted with a constant motor speed of 33 rpm on both sides of the motors. We found that the octopus robot swims fastest with stroke time ratio of 2.0:1. We illustrate its swimming performance of this ratio as in Fig. 6.

We measured and tracked the movement distance of the octopus robot and differentiated it to get velocity as a function of time. The experiment results are shown in Fig. 7. Fig. 7 (a) shows that swimming at a stroke time ratio of 2.0:1 is faster than at 1.6:1. Additionally, the 2.0:1 ratio, which closely resembles that of an real octopus, resulted in a completely positive swimming speed, indicating that the octopus did not swim backward. In both the 1.6:1 and 1.2:1 cases, the octopus swam backward. During the experiment, we found that the octopus robot with a 1.2:1 stroke time ratio had a different motion cycle than the other two, although repeatability is good. This is irregular. We analyzed the experiment results and concluded that this is because the motor is critically overloaded in that case.

\begin{figure}[H]
\centerline{\includegraphics[width=0.75\columnwidth]{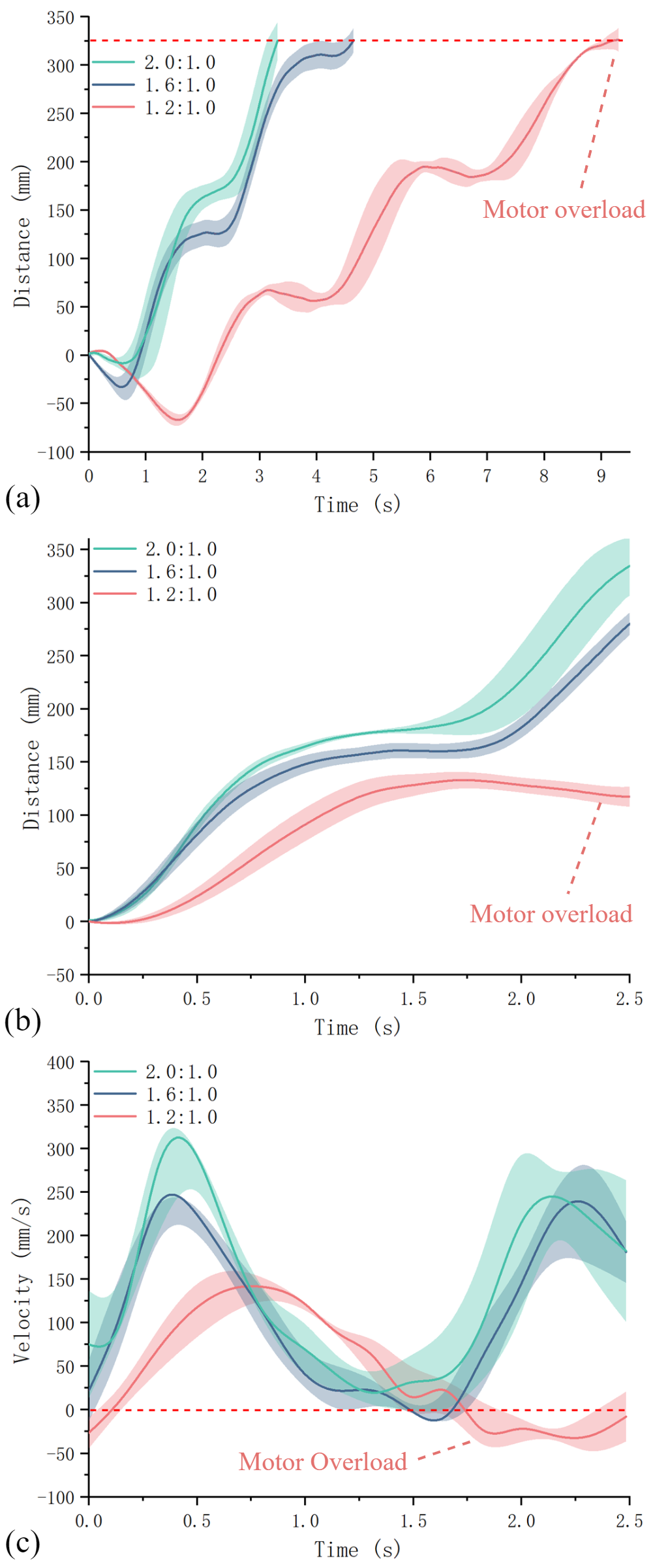}}
\vspace{-2mm}
\caption{Comparison of swimming performance of mechanisms with different stroke time ratios. a) The swimming distance of the octopus robot as a function of time. b) The aligned swimming distance according to their first stable stroke period. c) The velocity of the octopus robot as a function of time.}
\label{Fig. 7}
\vspace{-7mm}
\end{figure}

\section{Discussion}

\subsection{Different Incision Depth Octopus Arms Trajectory Analysis}

The octopus soft arms were tested with different incision depths by constant-speed rotation motor drive. We manually tracked the midline of the arm as a movement trajectory to compare their similarity with the biological finding (Fig. 5).

Due to the unique asymmetrical arm design, the arm exhibited different morphing when moving up and down. The arm tended to more straighten when pushing water downward, demonstrating greater stiffness. However, during the upward recovery motion, the arm bent significantly, which nicely illustrates its flexibility to interact with the water drag and adapt to it passively. In addition to analyzing the overall curve, we also focused on the recurve at the end of the curve. We believe this is a unique and advantageous phenomenon, which also appears in the movement of biological octopus arms, as such a form indicates less water resistance when lifting the arm, reducing unwanted backward propulsion.
 
Taking the experimental trajectory produced by the arm with a 70\% cutting depth as an example, comparing it with the biological one (Fig. 5 (b)), it is evident that both the initial characteristic recurve and the trend of maximum curvature change are quite consistent. The distribution of similar colour curves and the sequential trend of colour changes also correspond. This indicates that our 70\% artificial arm closely resembles the shape and movement characteristics of octopus swimming.

By altering the cutting depth as shown from Fig. 5 (c) to (f), we designed different types of octopus arms to explore the impact of cutting depth on their movement behavior. We conducted repeated experiments on arms of varying cutting depths under the same motor speed settings. We completed the figures of arm position trajectory over time after averaging. According to the statistics, the early recurve phenomenon becomes less noticeable as the cutting depth decreases and disappears entirely when the cutting depth is zero. Therefore, whether analyzing from the feature recurve or the overall trend, we found that the movement pattern formed by the arm with a 70\% cutting degree is closest to that of a biological octopus.

\subsection{Velocity Comparison Analysis}

Fig. 7 (a) shows the overall displacement time diagram of the mechanism swimming process in water with different stroke time ratios. The connecting rod length of the mechanism with a time ratio of 1.2:1 is much longer than the other two, which limits the mechanism's output \cite{b17} and causes motor overload. As a result, the following analysis mainly discusses the data of the other two mechanisms. With a stroke time ratio of 2.0:1, the octopus robot moved from the bottom of the tank to the surface in 3.3 s. The mechanism with a stroke time ratio of 1.6:1 was not as fast as the 2.0:1 stroke ratio robot but still completed the move in 4.7 s. The final travel distance of both is 325 mm. Both kinds of octopus robots with different stroke time ratios completed the underwater swimming work well. 

The swimming cycle of this octopus robot commences when all its arms initiate power strokes. To mitigate the impact of the initial unstable movement on the final outcome, the motion data of the three mechanisms within 2.5 seconds from the new stroke starting period were aligned. The resulting displacement time diagram is shown in Fig. 7 (b). The displacement data of the octopus robot with three mechanisms all show periodic changes. When the slope of the displacement curve is high, the octopus robot is in the power stroke stage, and when the slope is low, it is in the recovery stroke stage. In one cycle, the octopus with three different stroke time ratios shifted 176 mm, 153 mm, and 128 mm. The stroke time ratio of the two mechanisms at 2.0:1 and 1.6:1 in one cycle reached an impressive average speed of 110 mm/s and 96 mm/s. Compared with the current average speed of 38 mm/s whole soft octopus robot \cite{b9}, we achieved better performances. 

The octopus robot based on the new soft fluid actuator reached a peak speed of 90 mm/s \cite{b9}, and the octopus robot containing SMA muscle filament reached a peak speed of 40 mm/s \cite{b10}. Fig. 7 (c) shows the speed variation of the three mechanisms over time. It shows that the green line is above the blue line almost all the time, which proves that the mechanism with a 2.0:1 ratio of mechanism time has the best swimming performance. Its peak speed can reach 314 mm/s, which is much larger than that of the octopus robots based on the new soft fluid actuator or the SMA muscle filament. This further successfully illustrates the potential capability of this design in octopus swimming robots field. 

Beyond this, the motor number is down to two compared to Kazakidi's proposed eight-arm octopus swimming robot with eight motors \cite{b12}. Our design also avoids using the complex and tricky SMA actuators instead with common waterproof motors. All this dramatically simplifies the complexity of the octopus robot control-action system. This design provides a very simplified but efficient swimming platform for the field of bionic octopus robots. Also, it offers more research possibilities for future robot optimization and function development.

\subsection{Steering Capability}

The octopus robot contains two pairs of driving mechanisms, each of them drives four artificial arms. This feature allows the robot to change direction as it swims through the water by differential actuation control. Fig. 8 shows the steerable swimming capability by input different variable speed profiles of the two motors. The speed of the motors is relatively slow due to the size limitation of the water tank. First, the octopus robot was set as static in the initial position (t = 0 s). After the movement began, the motor speed of the left arms was faster than the right ones. After 10 seconds, the position of the octopus robot is shown, shifted to the left while moving up. Subsequently, the right motor speed would be adjusted faster than the left motor. The position of the octopus robot changed again, moving to the right (t = 30 s). The position change demonstrates that the octopus swimming robot is capable of achieving steering in the water tank.

\begin{figure}[t!]
\centerline{\includegraphics[width=0.9\columnwidth]{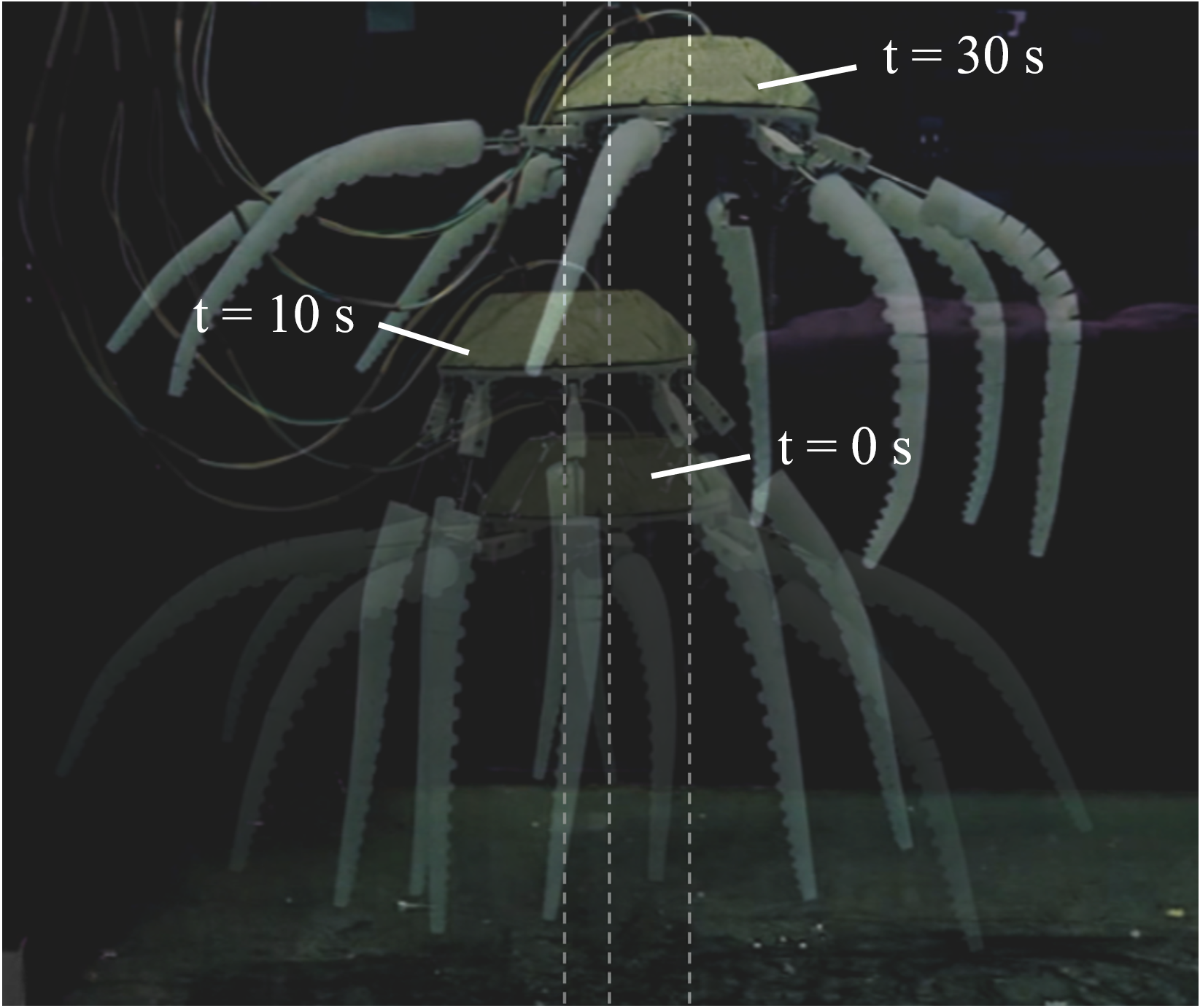}}
\vspace{-2mm}
\caption{The octopus robot shows steerable capability since its initial position at t = 0 s.}
\label{Fig. 8}
\vspace{-5mm}
\end{figure}

\section{Conclusion}
\balance

In this paper, we design a bionic octopus swimming robot with soft asymmetric stiffness octopus arms, with a quick-return mechanism which can freely adjust the time ratio between recovery stroke and power stroke. Experiments have shown that the bionic octopus arm with 70\% incision has the most similar movement trajectory to the biological octopus arm. After comparing the speed between different mechanisms, it is proved that this octopus robot has excellent underwater swimming ability, and the robot with a 2.0:1 stroke time ratio (similar to the biological octopus) swims the fastest in the tank. On the second power stroke, it reached an average speed of 110 mm/s and a peak speed of 314 mm/s.

In addition, the robot verifies the robot can achieve planar steerable swimming by using only two motors. It dramatically reduces the complexity of the actuation system while maintaining good swimming performance. This project presents a more accessible and efficient platform for biologists and roboticists, contributing to the field of bionic robotics at a lower cost. Furthermore, it may influence future design and applications in underwater vehicles.

\vspace{12pt}

\end{document}